\documentclass[conference]{IEEEtran}
\IEEEoverridecommandlockouts
\usepackage{cite}
\usepackage{amsmath,amssymb,amsfonts}
\usepackage{algorithmic}
\usepackage{graphicx}
\usepackage{textcomp}
\usepackage{xcolor}
\usepackage{booktabs}

\usepackage{booktabs}
\usepackage{mathrsfs}
\usepackage{textcomp, graphicx, subfigure}
\usepackage{epsfig}
\usepackage{amssymb}
\usepackage{amsmath}
\usepackage{amsfonts}
\usepackage{algorithmic}
\usepackage{algorithm}
\usepackage{cite}
\usepackage{mathtools,bbm}
\usepackage{enumerate}
\usepackage{lipsum}
\usepackage{mathtools}
\usepackage{cuted}
\usepackage{cancel}
\usepackage{verbatim} 
\usepackage{mathtools}
\usepackage{xcolor}
\usepackage{enumitem}
\usepackage{caption}

\usepackage[normalem]{ulem}
\usepackage{hyperref}

\hypersetup{
  citebordercolor=blue,
  filebordercolor=blue,
  linkbordercolor=blue
}

\makeatletter
\DeclareUrlCommand\ULurl@@{%
  \def\UrlLeft{\uline\bgroup}%
  \def\UrlRight{\egroup}}
\def\ULurl@#1{\hyper@linkurl{\ULurl@@{#1}}{#1}}
\DeclareRobustCommand*\ULurl{\hyper@normalise\ULurl@}
\makeatother

\def\BibTeX{{\rm B\kern-.05em{\sc i\kern-.025em b}\kern-.08em
    T\kern-.1667em\lower.7ex\hbox{E}\kern-.125emX}}
    
\begin{document}

\title{HyperSF: Spectral Hypergraph Coarsening via Flow-based Local  Clustering\\
 }



\author{\IEEEauthorblockN{ Ali Aghdaei }
\IEEEauthorblockA{
{Stevens Institute of Technology}\\
aaghdae1@stevens.edu }
\and
\IEEEauthorblockN{Zhiqiang Zhao}
\IEEEauthorblockA{
{Stevens Institute of Technology}\\
zzhao76@stevens.edu }
\and
\IEEEauthorblockN{Zhuo Feng}
\IEEEauthorblockA{
{Stevens Institute of Technology}\\
zhuo.feng@stevens.edu}

 }

\maketitle

\begin{abstract}
Hypergraphs allow modeling  problems with multi-way high-order relationships. However, the computational cost of most existing hypergraph-based algorithms can be heavily dependent upon the input hypergraph  sizes. To address the ever-increasing computational challenges, graph coarsening  can be potentially applied for preprocessing a given hypergraph by aggressively aggregating its vertices (nodes). However, state-of-the-art hypergraph partitioning (clustering) methods that incorporate heuristic  graph coarsening techniques  are not optimized for  preserving the  structural (global) properties of hypergraphs. In this work, we propose  an efficient spectral hypergraph coarsening scheme (HyperSF) for well preserving the original spectral (structural) properties of  hypergraphs.  Our approach  leverages a recent strongly-local max-flow-based clustering algorithm for detecting the sets of hypergraph vertices that minimize ratio cut.  To further improve the algorithm efficiency,  we propose a divide-and-conquer scheme by leveraging spectral clustering of the bipartite  graphs corresponding to the original hypergraphs. Our experimental results for a variety of hypergraphs extracted from real-world VLSI design benchmarks show that the proposed hypergraph coarsening algorithm can significantly improve the multi-way conductance of hypergraph clustering   as well as runtime efficiency when compared with existing state-of-the-art algorithms.  

\end{abstract}


\section{Introduction}
With an ascending trend in network size, it is indispensable to develop highly-scalable methods for boosting the performance of graph-related computations. To this end, graph coarsening is becoming a fundamental task in many graph-related tasks,  aiming to coarsen a given graph into a  much smaller one while preserving its essential structural properties \cite{jin2020graph, fahrbach2020faster, auer2012graph}.  Indeed, many tasks related to very-large-scale integration (VLSI) computer-aided design (CAD), numerical optimization, community detection, and graph clustering   have already benefited from  graph coarsening for improving  solution quality and runtime efficiency \cite{safro2015advanced, bianchi2020spectral, kahng2011vlsi, liu2019overlapping}.

Several different types of graph coarsening techniques have been proposed in the past decades. For example, heavy edge matching based graph coarsening method, such as Metis \cite{karypis1998fast}, has been widely used for graph partitioning; another popular way for graph coarsening is based on algebraic multigrid (AMG) inspired schemes \cite{livne2012lean, ron2011relaxation}, which   usually forms the Galerkin operator for generating the coarsened graphs;  \cite{ma2019graph} recently introduces a  graph neural network (GNN) based  framework to learn the  edge weights of the coarsened graphs.  Among   existing coarsening methods, spectral graph coarsening has been proven to be highly effective due to the preservation of key graph spectral (structural) properties \cite{loukas2018spectrally, loukas2019graph}. A variety of spectral graph coarsening schemes have been proposed in recent years: \cite{dorfler2012kron} proposed a Kron reduction  scheme based on   Schur complement; Purohit et al.  \cite{purohit2014fast} introduced   CoarseNet   to coarsen graphs while preserving the largest eigenvalue of its adjacency matrix such that the diffusion characteristics of the original graph can be kept; Loukas and Vandergheynst \cite{loukas2019graph} proposed a theoretical framework  based on   restricted spectral similarity, which is a modification of the previous spectral similarity metric for spectral graph sparsification; Bravo-Hermsdorff and Gunderson \cite{bravo2019unifying} proposed a probabilistic framework for graph coarsening, with the goal of preserving the inverse Laplacian of the coarsened graph; \cite{zhao2021towards} introduces a spectral  coarsening and scaling algorithm for preserving the first few eigenvalues and eigenvectors of the original graph Laplacian matrix.

Unlike simple graphs in which each edge only connects to two vertices, hypergraphs are a more versatile format for encapsulating  multi-relational   features.  As a result, hypergraph representation allows truthfully modeling the higher-order relationships, whereas simple graphs may fail to retain such information. For example, hypergraphs are more suitable than simple graphs in applications related to VLSI placement, co-authorship representations, and metabolic reactions 
\cite{lin2020dreamplace, lee2020hypergraph}.

To coarsen a hypergraph, the state-of-the-art hypergraph partitioning algorithms, such as  Hmetis \cite{10.1145/266021.266273}, Zoltan \cite{1639359}, and Mondriaan \cite{10.1137/S0036144502409019} can be directly adopted. However, it is not clear if such coarsening schemes can properly preserve the global (structural) properties of the original hypergraphs. Although  it is possible to perform spectral  clustering  by converting hypergraphs into  simple graphs using clique or star expansions \cite{10.1145/1143844.1143847}, the multi-way relationships may not be precisely represented. Moreover, although recent research proposed new methods for constructing hypergraph Laplacians \cite{inproceedings, 10.1145/3178123}, there exist no efficient implementations suitable for tackling large-scale hypergraph problems \cite{chan2020generalizing}.

In this work, we propose a scalable hypergraph coarsening framework that allows dramatically reducing the size of hypergraphs (number of vertices) by aggregating strongly-coupled vertices through a localized flow-based clustering method. Our approach allows preserving the key structural (spectral) properties of the original hypergraphs, which thus will substantially expedite the numerical computations of existing hypergraph-based algorithms without sacrificing solution quality. The major contribution of this work has been summarized as follows:
\begin{itemize}
\item We propose an efficient approach (HyperSF) for scaling down the hypergraph size by aggregating strongly-coupled nodes in local neighborhoods without impacting the global structure of the hypergraph. 
\item A key component of our method is a local, directed graph based max-flow algorithm for selecting the sets of vertices that minimize the local conductance, which can dramatically improve the quality of hypergraph coarsening.
\item We introduce a divide-and-conquer scheme to significantly  improve the algorithm  scalability by  carefully   bounding the size of each max $s$-$t$ flow network that is constructed using a set of hyperedges (node clusters) based on   spectral graph embeddings.
\item We have conducted extensive experiments  for  a variety of hypergraph test cases extracted from realistic VLSI  design problems, and obtained promising results when compared with existing hypergraph coarsening methods.
\end{itemize}
The rest of the paper is organized as follows. In Sections \ref{sec:sota} and \ref{sec:background}, we discuss the related works and provide a background introduction to the proposed method. In Section \ref{sec:method}, we present the proposed HyperSF method with detailed technical descriptions and algorithm flows. In Section \ref{sec:result}, we demonstrate extensive experimental results for a variety of real-word VLSI design benchmarks, which is followed by the conclusion of this work in Section \ref{sec:conclusion}.

\section{Related Works}\label{sec:sota}
Existing  hypergraph coarsening methods are based  on either vertex similarity or edge similarity \cite{Shaydulin_2019}: the edge similarity based coarsening techniques  contract  similar hyperedges with large sizes into smaller ones that include only a few vertices, which can be easily implemented but may impact the original hypergraph structural properties during the clustering process; the vertex-similarity based algorithms rely on checking the distances between  vertices for discovering  strongly-coupled (correlated)  clusters, which can be achieved  leveraging  hypergraph embedding  that maps each vertex into a  low-dimensional vector such that the Euclidean distance (coupling) between the vertices can be easily computed in constant time.

\subsection{Heuristic methods}
The state-of-the-art hypergraph coarsening algorithms are heuristic. For example, Hmetis, Zoltan, and Mondriaan utilize a coarsening method incorporating a multi-level partitioner to greedily bisect the hypergraph effectively \cite{10.1145/266021.266273, 1639359, 10.1137/S0036144502409019}. They leverage hyperedge \textit{inner product}  for contracting hyperedges that share more vertices; another popular hypergraph partitioner  PaToH \cite{ccatalyurek2011patoh},  uses the \textit{absorption matching} metric in its coarsening strategy. However, these simple metrics  do not capture the higher-order relationships in the hypergraph.

An \textit{algebraic distance} criteria is introduced to improve the previous works\cite{Shaydulin_2019}, while a relaxation-based method is adopted for assigning a coordinate to each vertex and subsequently computing the Euclidean distance between the maximally-distanced vertices within a hyperedge. Then  the hyperedge weights are updated accordingly and integrated into the Zoltan partitioner for improving hypergraph partitioning. However, such a method does not immediately lead to a spectral hypergraph coarsening framework since many heuristics have been used for minimizing hypergraph cuts.

\subsection{Spectral methods}
Spectral methods provide a formidable methodology for the theoretical computer science applications. Researchers extensively studied spectral graph coarsening to develop high-performance algorithms with lower complexity. 

Prior work generalized the existing spectral graph coarsening methods for hypergraphs by converting the hyperedges into simple graph edges using star or clique expansion \cite{159993}. However, these methods may result in lower performance due to ignoring the multi-way relationship between the entities. A more rigorous approach by Tasuku and Yuichi \cite{soma2018spectral} generalized  spectral graph sparsification for hypergraph setting by sampling the hyperedges according to the probability determined based on the ratio of the hyperedge weight over the minimum degree of two vertices inside the hyperedge.

Another family of spectral methods for hypergraphs explicitly builds the Laplacian matrix to analyze the critical properties of hypergraphs \cite{fu2019hplapgcn}: Zhou et al. propose a method to create the Laplacian matrix of a hypergraph and generalize graph learning algorithms for hypergraph applications \cite{inproceedings}; Chan et al. leverage the diffusion process to introduce a hypergraph Laplacian operator by measuring the flow distribution within each hyperedge \cite{10.1145/3178123}; later, they present a mediator-based diffusion algorithm to provide a well-approximated non-linear quadratic formula for hypergraphs \cite{chan2020generalizing}. However, these methods are only based on theoretical analysis, which do not allow for  practically-efficient implementations.

\subsection{Flow-based methods}
Many graph-related   algorithms extensively exploit flow-based techniques for various purposes. In \cite{yang2003efficient}, a graph partitioner is proposed to split a graph into the balanced partitions by minimizing the min-cut objective.
Additionally, several graph clustering methods find strongly coupled vertices in a local neighborhood by solving a max $s$-$t$ flow, min $s$-$t$ cut problem for a subgroup of entities to find clusters that minimize the ratio cut \cite{orecchia2014flow, satuluri2009scalable}. Such algorithms guarantee the solution quality with a reasonably fast runtime. For example,  the algorithm introduced in \cite{veldt2019flow} accepts a small ratio of the target cluster as the seed nodes and extends the network around them to solve the max $s$-$t$ flow, min $s$-$t$ cut problem locally, which subsequently clusters the vertices that minimize the localized conductance. 

\section{Background}\label{sec:background}
\subsection{Spectral graph theory for undirected graphs}

For an undirected graph $G=(V,E,w)$, $V$ denotes a set of nodes (vertices), $E$ denotes a set of (undirected) edges, and $w$ denotes the associated edge weights. 
We define ${D}$ to be a diagonal matrix with ${D}(i,i)$ being equal to the (weighted)  degree of node $i$, and ${A}$ to be the adjacency matrix of undirected graph $G$ as follows:
\begin{equation}\label{di_adjacency}
{A}(i,j)=\begin{cases}
w(i,j) & \text{ if } (i,j)\in E \\
0 & \text{otherwise }.
\end{cases}
\end{equation}
Then, the Laplacian matrix of the graph $G$ can be calculated by $L = D-A$, which satisfies the following conditions: (1) The sum of each column or row equals zero; (2) All off-diagonal elements are non-positive; (3) The graph Laplacian is a symmetric diagonally dominant (SDD) matrix with non-negative eigenvalues.

\noindent(\textit{Courant-Fischer Minimax Theorem}) The $k$-th largest eigenvalue of the Laplacian matrix $L \in \mathbb{R}^{|V|\times|V|}$ can be computed as follows:
\begin{equation}\label{eqn:minmax}
    \lambda_k(L) = \min_{dim(U)=k}\,{\max_{\substack{x \in U \\ x \neq 0}}{\frac{x^\top Lx}{x^\top x}}},
\end{equation}
which can be leveraged for computing the spectrum of the Laplacian matrix $L$. Given a graph $G=(V,E,w)$ with the vertices partitioned into $(S, \hat{S})$, the conductance of the partition $S$ is defined as follows:
\begin{equation}\label{eqn:s_conductance}
    \Phi_G(S) = \frac{w(S, \hat{S})}{\min\left(vol(S), vol(\hat{S})\right)} = \frac{\sum_{(i,j)\in E:i\in S,j\notin S}{w(i, j)}}{\min\left(vol(S), vol(\hat{S})\right)},
\end{equation}
where the volume of the partition $vol(S)$ is defined as the sum of the (weighted) degree of vertices in     partition $S$, which can be denoted as follows:
\begin{equation}\label{eqn:partition_volume}
    vol(S) := \sum_{i \in S}{d(i)}.
\end{equation}
The conductance of the graph \cite{chung1997spectral} $G$ can be  defined as:
\begin{equation}\label{eqn:G_conductance}
    \Phi(G)= \min_{vol(S) \leq vol(V)/2}{\Phi_G(S)}.
\end{equation}
It has been shown that the graph conductance $\Phi(G)$ is closely related to the spectral property of its graph $G$, which can be revealed by the \textit{Cheeger's inequality} \cite{chung1997spectral} as follows:
\begin{equation}\label{eqn:cheeger}
    \omega_2/2 \leq \Phi(G) \leq \sqrt{2\omega_2},
\end{equation}
where $\omega_2$ is the 2nd smallest eigenvalue of the normalized Laplacian matrix $\widetilde{L}$ defined as $\widetilde{L} = D^{-1/2}LD^{-1/2}$.

Based on the definition of the graph conductance, local conductance concept has been utilized for graph partitioning \cite{veldt2020minimizing}. Given a reference node set $R \subseteq V$, the local conductance regarding to the node set $R$ can be defined as  \cite{veldt2020minimizing}:
\begin{equation}\label{eqn:s_conductance}
    \Phi_R(S) = \frac{\sum_{(i,j)\in E:i\in S,j\notin S}{w(i, j)}}{vol(S\cap R)-\delta vol(S\cap \hat{R})},
\end{equation}
where $\delta$ is a locality parameter which controls the penalty for including nearby nodes outside set $R$.


\subsection{Spectral graph theory for hypergraphs}
We denote a hypergraph $H = (V, E)$, where $V$ is the set of vertices and $E$ is the set of hyperedges with unit weight. We specify $n:= \left| V \right|$ and $m:= \left| E \right|$ to be the numbers of vertices and hyperedges, respectively. We define a vertex degree to be: 
\begin{equation}
    d_v := \Sigma_{e \in E : v \in e} w(e),
\end{equation}
where $w(e)$ is the hyperedge weight. We define the hypergraph volume for node set $S \subseteq V$ to be:
\begin{equation}
    vol(S) := \Sigma_{v \in S}d_v.
\end{equation}
Then, the  conductance of a given set $S$ is defined as:
\begin{equation}
    \Phi(S) := \frac{cut(S, \hat{S})}{min\{vol(S), vol(\hat{S})\}},
\end{equation}
where $cut(S, \hat{S})$ is the number of crossing hyperedges between $S$ and $\hat{S}$.  We use \textit{all or nothing} splitting function to compute the \textit{cut} that is penalizing the hyperedges the same way regardless of how splitting them. The hypergraph conductance is defined as $\Phi_H := \min\limits_{\emptyset \nsubseteq S \subseteq V} \Phi(S)$. Nate et al. introduce hypergraph local conductance (HLC) as \cite{veldt2020minimizing}: 
\begin{equation}\label{obj:HLC}
    \textbf{HLC}_R(S) = \frac{cut(S, \hat{S})}{vol(S \cap R) - \delta vol(S \cap \hat{R})},
\end{equation}
where   $R$ is the set of seed vertices  given as the input.
\section{Spectral Hypergraph Coarsening}\label{sec:method}
\subsection{Overview of our method}
Let $H = (V, E)$ be the given hypergraph dataset; our coarsening algorithm aims to generate $H_s = (V_s, E_s)$, where $H_s$ is spectrally-similar to $H$. We create a spectral coarsening scheme that consists of the following steps:

\begin{itemize}
    \item \textbf{Step A} will produce an initial set of node clusters within each hyperedge leveraging undirected graph embedding.
    \item \textbf{Step B} will aggregate spectrally-similar vertices in a local neighborhood  using a flow-based clustering method \cite{veldt2020minimizing}. 
\end{itemize}

\begin{figure*}[h]
    \centering
    \includegraphics [ width=0.915028\textwidth]{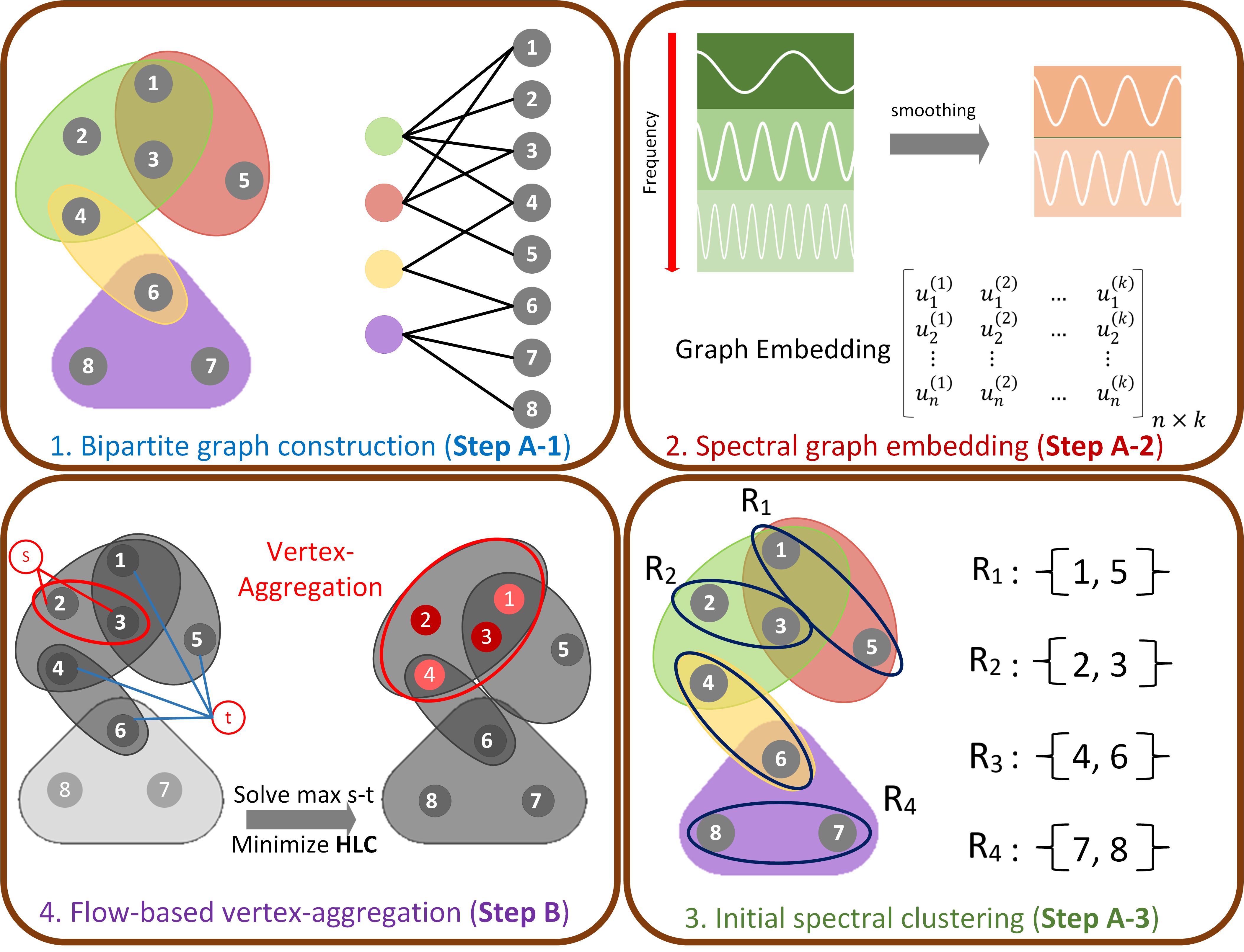}
    \caption{Overview of the proposed HyperSF framework}
    \label{fig:overview}
\end{figure*}

Fig.~\ref{fig:overview} shows the proposed HyperSF framework, which consists of four key phases: phase (1) constructs the bipartite graph based on the original hypergraph; phase (2) maps the vertices into a low-dimensional space leveraging spectral graph embedding; phase (3) applies spectral graph clustering to split the vertices within each hyperedge to create the initial node clusters; phase (4) solves the max $s$-$t$ flow, min $s$-$t$ cut problem on a directed graph converted from local hyperedges to determine the final set of nodes for aggregation.

\subsection{\textbf{Step A:} Initial spectral graph clustering}
We apply star expansion to model the hypergraph with a pairwise relationship between the vertices so that the eigenvectors of the graph Laplacian matrix can be leveraged for spectral graph embedding (vector representation of  vertices). 
Since we want to cluster highly related vertices in a local neighborhood so that the coarsened graph will preserve the structural properties of the original hypergraph. This requires to approximate the first few (low-frequency) eigenvectors related to the key spectral properties of the graph. To achieve this goal, we leverage the following scalable  low-pass filtering algorithm to effectively remove high-frequency graph signal components and subsequently map the each vertex into a $k$-dimensional vector, where $k$ denotes the number of smoothed  testing vectors (graph signals).

Let $u$ be a random vector in which each element corresponds to a  vertex in the graph. $u$ can be considered as a linear combination of many eigenvectors. Existing iterative methods like Gauss-Seidel and Jacobi relaxation procedures can be leveraged as low-pass graph filters for removing  highly oscillating components in $u$, which will produce smoothed vectors that correspond to the linear combination of the first few eigenvectors:
\begin{equation}
    u = \Sigma_{i=1}^n \gamma_i x_i  \Rightarrow \hat{u} = \Sigma_{i=1}^{\hat{n}} \hat{\gamma_i} x_i  \quad \hat{n} \ll n, 
\end{equation}
where $\gamma_i$ and $\hat{\gamma_i}$ are the weighting coefficients, and $\hat{u}$ is the smoothed vector that is obtained after applying a few steps of the Gauss-Seidel iterations on $u$. Applying the above filtering (smoothing) function for $k$ different random vectors (orthogonal to the all-one vector) will result in $k$  smoothed vectors $K = (u^{(1)} ... u^{(k)})$ that allow embedding the graph into a $k$-dimensional space.  Next, we repeatedly apply  spectral clustering for each hyperedge (sorted with descending cardinalities) to find node clusters based on their embedding vectors (smoothed vectors) and the hypergraph coarsening (reduction) ratio. When processing each hyperedge, it is important to flag the nodes that have  already been processed, and skip the following hyperedges that include the flagged nodes.  Algorithm \ref{alg:graph-based} provides the details of the proposed initial spectral graph clustering scheme (Step A-3 shown in Fig. \ref{fig:overview}). 

\begin{algorithm}
\small { \caption{Initial spectral graph clustering}\label{alg:graph-based}}
\textbf{Input:} Hypergraph $H = (V,E)$, $n =\left| V\right|$, $m = \left| E\right|$, and embedding dimension $k$;\\
\textbf{Output:} {A set of vertex clusters $\mathcal{R}$};\\
  \algsetup{indent=1em, linenosize=\small} \algsetup{indent=1em}
    \begin{algorithmic}[1]
    \STATE{Construct the bipartite graph $G$ corresponding to the hypergraph $H$ by applying the star model };
    \STATE{Generate $k$ different random vector related to $G$};
    \STATE{Perform smoothing function to remove the high frequency components};
    \STATE{Map the graph vertices into a $k$-dimensional space using the smoothed vectors};
    \STATE{Sort the hyperedges according to their cardinality: 
    \\ $|e_1| > |e_2| > ... > |e_m|$};
    \FOR{$i \gets 1$ to $m$}
     \STATE $R^{(i)} \gets$ cluster the vertices in $e_i$ using $k$-means method;
     \ENDFOR
     \STATE Return $\mathcal{R}$ .
    \end{algorithmic}
\end{algorithm}

\subsection{\textbf{Step B: } Flow-based hypergraph local clustering}
In this section, we improve the quality of the initial spectral graph clustering by incorporating a flow-based low clustering technique \cite{veldt2020minimizing} for aggregating vertices with minimal impact on the global hypergraph structure. 

\subsubsection{Flow-based  hypergraph clustering}
Semi-supervised clustering methods utilize flow-based techniques to find a cluster of vertices strongly connected to the seed nodes $R$, which repeatedly solve max $s$-$t$ flow, min $s$-$t$ cut problem  to minimize \textbf{HLC}.  

\begin{figure}[]
    \centering
    \includegraphics [  width=0.45\textwidth]{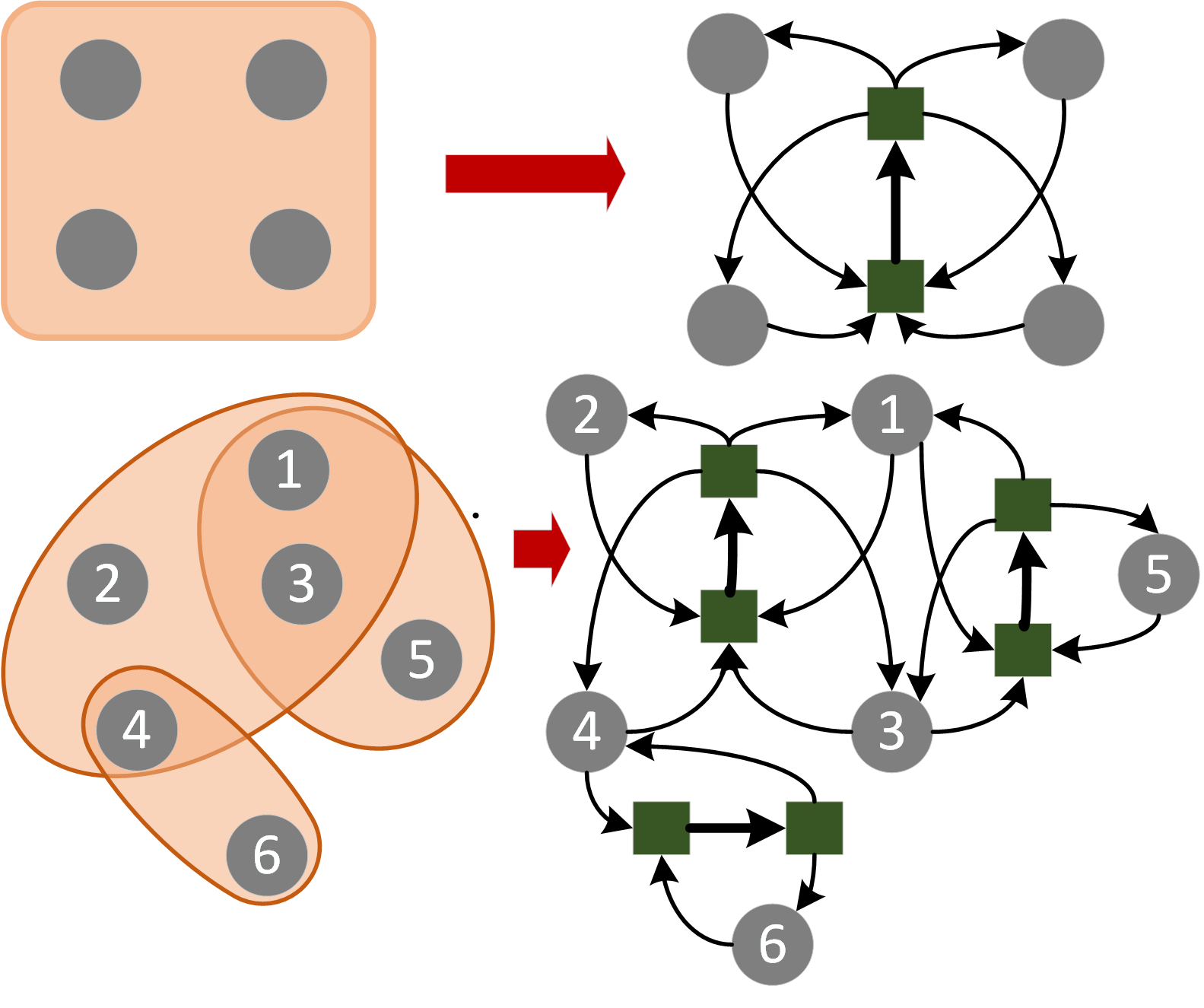}
    \caption{Hypergraph to directed graph conversion}
    \label{fig:gadget}
\end{figure}

We formalize the hypergraph coarsening problem into a semi-supervised clustering task to aggregate the strongly-connected vertices that allow minimizing (\ref{obj:HLC}). To this end, the flow-based clustering method utilizes the output of the initial spectral graph clustering (Algorithm \ref{alg:graph-based}) and treats each cluster as a seed node set $R$. For each seed node set, we  repeatedly solve a max $s$-$t$ flow, min $s$-$t$ cut problem to decide a set of node clusters   that minimizes the localized conductance. To this end,   we first create the auxiliary hypergraph by introducing a source vertex $s$ and sink vertex $t$.  Next, we replace each hyperedge with a directed graph by creating a network: for each vertex $r \in R$ we introduce an edge $(s, r)$, whereas  for each $j \in \hat{R}$ we introduce an edge $(j, t)$; we also introduce two auxiliary vertices (e.g. the green squares shown in Fig. \ref{fig:gadget}) $a$ and $a'$ and create a directed edge from $a$ to $a'$.  In the last, for each vertex $\{v \in e \}_{e \in E}$, we introduce a directed edge $(a, v)$ and  a directed edge  $(v, a')$. We show how to build the directed graph model for a given set of hyperedges in Fig. \ref{fig:gadget}.

We initialize $S = R$ and iteratively update $S$ to minimize \textbf{HLC} by repeatedly solving the  max $s$-$t$ flow, min $s$-$t$ cut problem on the created auxiliary hypergraph to minimize the hypergraph cut \cite{veldt2020hypergraph}:
\begin{equation}\label{eq:cutH}
    cut^{s-t}(S) = cut_H(S) + vol_H(\hat{S} \cap R) + \delta vol_H(S \cap \hat{R}).
\end{equation}
In the last, we aggregate the output set of vertices obtained from flow-based methods that minimize the localized conductance to produce a smaller hypergraph with fewer vertices while preserving the key structural properties of the original hypergraph. We flag the nodes that are already clustered to avoid assigning the same node to different clusters.

\subsubsection{Flow-based local clustering}
An algorithm is local if the input is a small portion of the original dataset. The aforementioned flow-based hypergraph clustering algorithm can be made strongly local when expanding the network around the seed nodes $R$, which will obviously  benefit the proposed hypergraph coarsening framework: (1) applying the max $s$-$t$ flow, min $s$-$t$ cut problem on the local neighborhood of the seed nodes restricts node-aggregation locally and keeps the global hypergraph structure intact; (2) such a local clustering scheme will significantly improve the algorithm efficiency due to the small-scale input dataset.


To achieve flow-based local   clustering of hypergraph nodes, we first construct a sub-hypergraph $H_L$ by iteratively expanding the hypergraph around the seed node set $R$ and then repeatedly solve the hypergraph cut problem to minimize \textbf{HLC} until no significant changes in local conductance are observed. Let $H = (V, E)$ be the hypergraph. We define $E(S) = \cup_{v \in V, v \in \{e\}_{e \in E}}E(v)$ for any set $S \subseteq V$. We denote $H_\varsigma = (V \cup \{s,t\}, E \cup E^{st})$, where $E^{st}$ is the terminal edge set. We aim to construct a sub-hypergraph $H_L$ to replace $H_\varsigma$ that minimizes \textbf{HLC} by repeatedly solving a local version of (\ref{eq:cutH}). To this end, we set up an oracle to discover a set of best neighborhood vertices for a given vertex $v$:
\begin{equation}
    \kappa (v) = \{u \in V\}_{(u,v)\in e, e\in E}.
\end{equation}
We let the oracle accept a set of seed nodes $R$ and return $\kappa (R) = \cup_{v \in R} \kappa (v)$. By utilizing the best neighborhood of the seed nodes $\kappa (R)$, we build a local hypergraph $H_L = (V_L \cup \{s,t\}, E_L \cup E_L^{st} )$, where $V_L = R \cup \kappa (R)$ and $E_L = \{e \in E \mid  V_L \in e\}$. We create the local auxiliary hypergraph of $H_L$ by introducing the source node $s$ and sink node $t$, so that $E_L^{st} \subseteq E^{st}$, and repeatedly solve the max $s$-$t$ flow, min $s$-$t$ cut problem to minimize \textbf{HLC}. The algorithm continuously expands   $H_L$ and includes more vertices and hyperedges from $H_\varsigma$ by solving (\ref{eq:cutH}) for the local hypergraph $H_L$.

\begin{algorithm}
\small { \caption{Flow-based hypergraph clustering}\label{alg:flow_based}}
\textbf{Input:} Hypergraph $H = (V,E)$, $R \subseteq \mathcal{R}$, and $\epsilon$;\\
\textbf{Output:} {A set of vertices $S$ that minimizes \textbf{HLC(S)}};\\
  \algsetup{indent=1em, linenosize=\small} \algsetup{indent=1em}
    \begin{algorithmic}[1]
    \STATE Assign the seed nodes $S \gets R$;  
    \STATE $\Delta_{\textbf{HLC}} \gets \infty$; 
    \WHILE{$\Delta_{\textbf{HLC}} > \epsilon$}
    \STATE Identify the best neighborhood of seed nodes $\kappa(S)$;
    \STATE Update $S$ according to $\kappa(S)$ to construct   $H_L$; 
    \STATE Add a source node $s$ and sink node $t$ to $H_L$;
    \STATE Repeatedly solve the max $s$-$t$ flow, min $s$-$t$ cut problem by minimizing (\ref{eq:cutH}) for $H_L$;
    \STATE $\Delta_{\textbf{HLC}} \gets \textbf{HLC(S)}^j - \textbf{HLC(S)}^{j-1}$;
    \ENDWHILE
     \STATE Return $S$. 
    \end{algorithmic}
\end{algorithm}

In Algorithm \ref{alg:flow_based} we present the details of the flow based local clustering technique. It accepts the original hypergraph $H=(V, E)$, a set of seed nodes in $R \subseteq \mathcal{R}$, and the convergence parameter $\epsilon$, which will output a set of strongly connected vertices $S$ that minimizes \textbf{HLC}.

\subsection{Divide-and-conquer}
To further improve the flow-based algorithm efficiency, we leverage a divide-and-conquer approach to partition the dataset into different parts with strongly connected vertices and perform the flow-based method separately in each partition. To this end, we use Metis as a standalone partitioning tool to split the undirected bipartite graph corresponding to the original hypergraph. The flow-based method produces the clusters in a parallel way that expedites the proposed node-aggregation procedure. Additionally, the number of partitions determines the runtime and quality of the algorithm. Our method allows a flexible trade-off between algorithm efficiency and solution accuracy, which can be achieved by adjusting the size of each partition. For example, choosing smaller partitions will result in a higher runtime efficiency but negatively impact hypergraph clustering (coarsening) solution. In Algorithm \ref{alg:aggregation}, we illustrate the entire procedure for spectral hypergraph coarsening.

\begin{algorithm}
\small { \caption{The HyperSF algorithm flow}\label{alg:aggregation}}
\textbf{Input:} Hypergraph $H = (V,E)$, $n =\left| V\right|$, $m = \left| E\right|$;\\
\textbf{Output:} {Coarsened hypergraph $H_c = (V_c, E)$, where $\left| V_c \right| \ll n$};\\
  \algsetup{indent=1em, linenosize=\small} \algsetup{indent=1em}
    \begin{algorithmic}[1]
    \STATE Call Algorithm \ref{alg:graph-based} to compute $\mathcal{R}$ via spectral embedding of bipartite graph $G$;
    \STATE $\{p\in V\}_{\subseteq P} \gets$ Apply Metis partitioning to obtain a set of vertices in the same region;
    \FOR{$j \gets 1$ to $\left|P \right|$}
    \STATE $\hat{R} \gets R_{\subseteq \mathcal{R}} \cap \{p \}_{\subseteq P}^j$;
    \FOR{$i \gets 1$ to $\left| \hat{R} \right|$}
    \STATE Choose a set of seed nodes $S \gets$ $\hat{R}^{i}$;
    \STATE $\hat{S} \gets$ Call Algorithm \ref{alg:flow_based} to reach a set of strongly connected vertices;
    \STATE Aggregate $\{v \in \hat{S}\}_{\hat{S} \subseteq V}$;
    \ENDFOR 
    \ENDFOR 
     \STATE Return  $H_c$, where $\left|V_c \right| \ll n$. 
    \end{algorithmic}
\end{algorithm}

\subsection{Algorithm complexity of HyperSF}
The   runtime complexity of \textbf{Step A} for the graph-based spectral clustering   is $O(E)$ since we only perform the node clustering within each hyperedge; the runtime complexity of  \textbf{Step B} is $O\left(k^3 vol_H(R)^3(1 + \epsilon ^ {-1})^3\right)$, where $k$ is the maximum hyperedge cardinality. Accordingly, the worst-case runtime complexity of the entire HyperSF algorithm  is $O\left(nk^3 vol_H(R)^3(1 + \epsilon ^ {-1})^3\right)$. 

\section{Experimental Results}\label{sec:result}
We conduct extensive experiments to analyze the performance of the proposed spectral hypergraph coarsening method using a set of VLSI-related datasets ``ibm01" to ``ibm18''  \footnote[1]{https://vlsicad.ucsd.edu/UCLAWeb/cheese/ispd98.html}. By testing other hypergraph coarsening techniques on the same data sets, we can provide a fair comparison among different approaches. We have implemented HyperSF in Julia by incorporating the graph spectral embedding technique with the local hypergraph flow-based clustering. All experiments ran on a laptop with 8 GB of RAM and a 2.2 GHz Quad-Core Intel Core i7 processor. An implementation of our algorithm and the code for reproducing our experimental results are available online at \ULurl{https://github.com/aghdaei/HyperSF}.

\subsection{The results of spectral hypergraph coarsening}
We compare the performance of the proposed spectral hypergraph coarsening method (HyperSF) with other hypergraph coarsening frameworks.  We compare our algorithm with non-spectral coarsening methods used in hypergraph partitioning methods to investigate our algorithm's performance thoroughly. {To the best of our knowledge,  the state-of-the-art hypergraph partitioners leverage  heuristic-based hypergraph coarsening frameworks to contract the hyperedges based on  edge similarity criteria.} In addition, we  also construct the bipartite graph of the original hypergraph and use the existing spectral methods to reduce the hypergraph size.
We used the average conductance of all clusters   to evaluate the performance of HyperSF when comparing with the other hypergraph coarsening algorithms. According to the \textit{Cheeger's inequality}, a smaller average conductance  implies a better graph coarsening solution.

To this end, we compute the following average local conductance $\Phi$ of the node clusters produced by each  method:
\begin{equation}
    \Phi = \frac{1}{\left| S\right|}\Sigma_{i = 1}^{\left| S \right|}\textbf{HLC}(S^i).
\end{equation}
\subsubsection{Spectral methods}
\begin{figure}[]
    \centering
    \includegraphics [  width=0.45\textwidth]{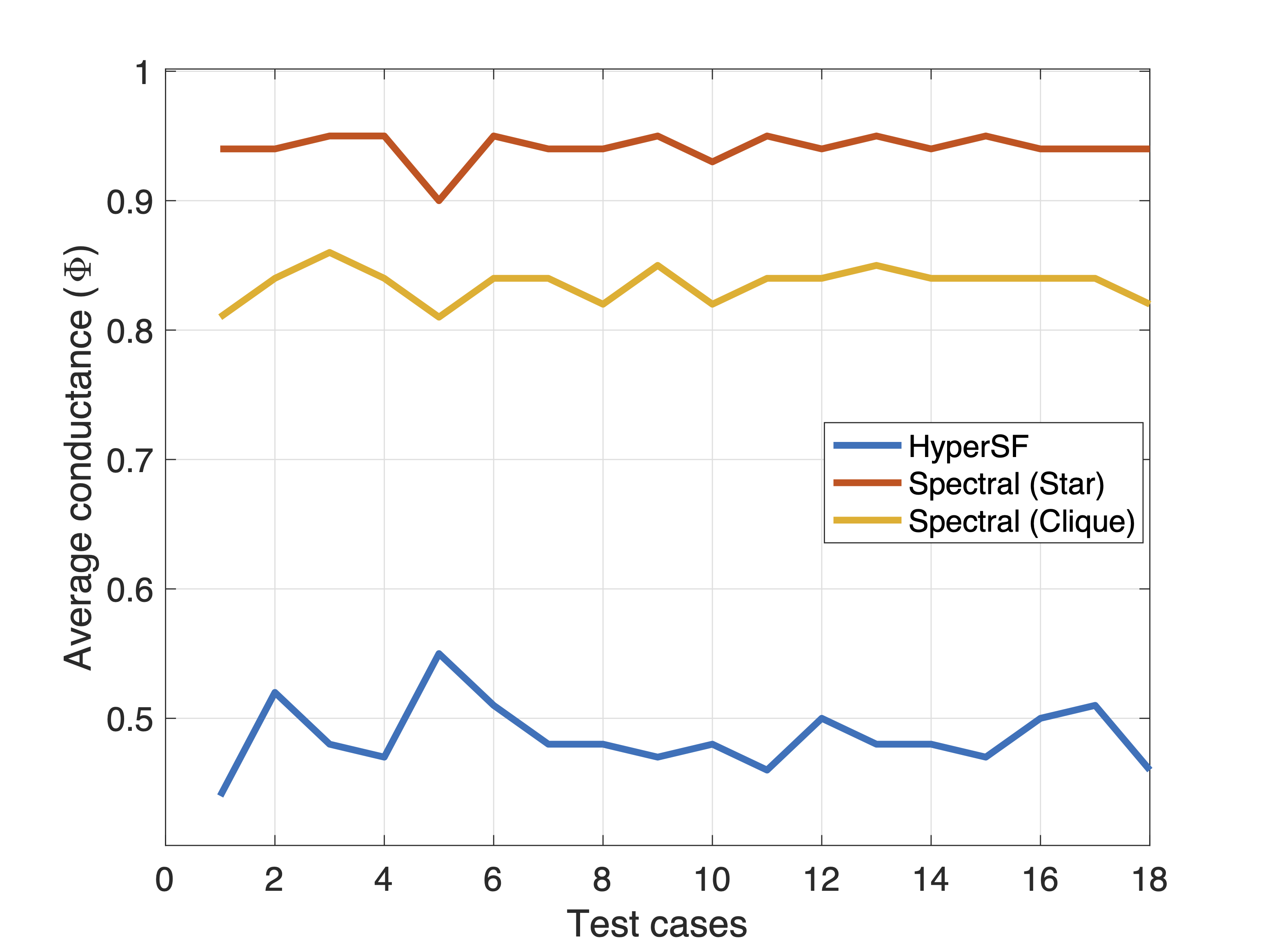}
    \caption{Average conductance ($\Phi$) comparison: HyperSF vs spectral methods using all ibm test cases}
    \label{fig:spectral}
\end{figure}

Although there is no existing spectral hypergraph coarsening framework to compare with HyperSF,  we  can apply star and clique models to construct the corresponding undirected bipartite graphs associated with the hypergraph. Once a simple graph is constructed, the traditional spectral graph coarsening methods  \cite{zhao2021towards} can be leveraged for aggregating the spectrally-similar vertices and generating a smaller undirected graph. Once the node clusters are produced, we can  compute \textbf{HLC} of each node set and thus $\Phi$ for the original hypergraph. Fig. \ref{fig:spectral} shows the computed average local conductance $\Phi$ of   spectral hypergraph coarsening methods for all   test sets using the same coarsening ratio. The results show that HyperSF always achieves the lowest $\Phi$ for all the datasets. 
\subsubsection{Non-spectral methods}
\begin{figure}[h]
    \centering
    \includegraphics [  width=0.45\textwidth]{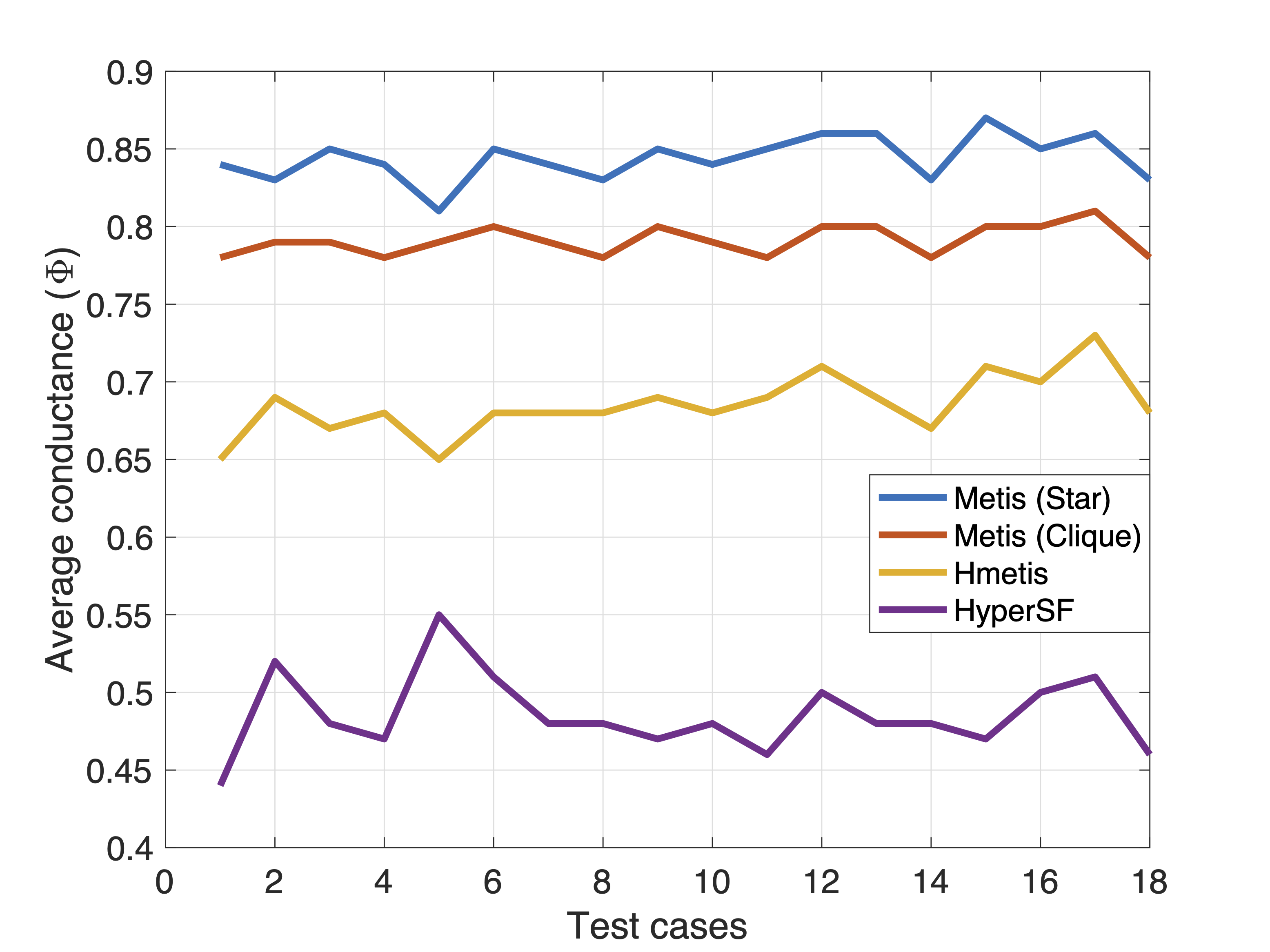}
    \caption{Average conductance ($\Phi$) comparison: HyperSF vs non-spectral methods using all ibm test cases}
    \label{fig:non-spectral}
\end{figure}

In this section, we compare the performance of HyperSF with two non-spectral hypergraph coarsening approaches. The first method directly applies Metis on the undirected graphs converted from the original hypergraph (using star and clique expansions) and aggregates the vertices within each partition to create a smaller  hypergraph. The second method utilizes a popular hypergraph partitioning tool, Hmetis, to aggregate strongly-connected vertices in each partition. To provide a fair comparison, we apply the same coarsening ratio in different methods  such that   all coarsened hypergraphs $H_s = (V_s, E_s)$ will have the same size $\left|V_s \right|$. In Fig. \ref{fig:non-spectral}, we show the performance of Metis (using star and clique models), Hmetis, and HyperSF by computing the average local conductance $\Phi$ for all  test cases. We observe that HyperSF always achieves  significantly better coarsening results.

\begin{table*}[h]
\centering
\caption{The average local conductance ($\Phi$) comparison between HyperSF and other coarsening methods}
\begin{tabular}{@{}cccccccccc@{}}
\toprule
Hypergraphs & V & E & RR (\%) & Spectral ($\mathcal{S}$) & Spectral ($\mathcal{C}$) & Metis ($\mathcal{S}$) & Metis ($\mathcal{C}$) & Hmetis & HyperSF \\ \midrule
ibm01  & 12752  & 14111  & 75 & 0.94 & 0.81 & 0.84 & 0.78 & 0.65 & 0.44 \\
ibm02  & 19601  & 19584  & 75 & 0.94 & 0.84 & 0.83 & 0.79 & 0.69 & 0.52 \\
ibm03  & 23136  & 27401  & 75 & 0.95 & 0.86 & 0.85 & 0.79 & 0.67 & 0.48 \\
ibm04  & 27507  & 31970  & 75 & 0.95 & 0.84 & 0.84 & 0.78 & 0.68 & 0.47 \\
ibm05  & 29347  & 28446  & 75 & 0.90 & 0.81 & 0.81 & 0.79 & 0.65 & 0.55 \\
ibm06  & 32498  & 34826  & 75 & 0.95 & 0.84 & 0.85 & 0.80 & 0.68 & 0.51 \\
ibm07  & 45926  & 48117  & 75 & 0.94 & 0.84 & 0.84 & 0.79 & 0.68 & 0.48 \\
ibm08  & 51309  & 50513  & 75 & 0.94 & 0.82 & 0.83 & 0.78 & 0.68 & 0.48 \\
ibm09  & 53395  & 60902  & 75 & 0.95 & 0.85 & 0.85 & 0.80 & 0.69 & 0.47 \\
ibm10  & 69429  & 75196  & 75 & 0.93 & 0.82 & 0.84 & 0.79 & 0.68 & 0.48 \\
ibm11  & 70558  & 81454  & 75 & 0.95 & 0.84 & 0.85 & 0.78 & 0.69 & 0.46 \\
ibm12  & 71076  & 77240  & 75 & 0.94 & 0.84 & 0.86 & 0.80 & 0.71 & 0.50 \\
ibm13  & 84199  & 99666  & 75 & 0.95 & 0.85 & 0.86 & 0.80 & 0.69 & 0.48 \\
ibm14  & 147605 & 152772 & 75 & 0.94 & 0.84 & 0.83 & 0.78 & 0.67 & 0.48 \\
ibm15  & 161570 & 186608 & 75 & 0.95 & 0.84 & 0.87 & 0.80 & 0.71 & 0.47 \\
ibm16  & 183484 & 190048 & 75 & 0.94 & 0.84 & 0.85 & 0.80 & 0.70 & 0.50 \\
ibm17  & 185495 & 189581 & 75 & 0.94 & 0.84 & 0.86 & 0.81 & 0.73 & 0.51 \\
ibm18 & 210613 & 201920 & 75 & 0.94 & 0.82 & 0.83 & 0.78 & 0.68 & 0.46 \\ 
\bottomrule
\end{tabular}
\label{tab:average conductance}
\end{table*}

TABLE \ref{tab:average conductance} shows the average local conductance $\Phi$ of different methods using the same hypergraph reduction ratio (RR). We reduce the number of nodes in each original hypergraph by $75\%$   (e.g., for a hypergraph with $100$ nodes,  the coarsened hypergraph will have only $25$ nodes). For simplicity, we denote the star and clique models by $\mathcal{S}$ and $\mathcal{C}$, respectively. The experimental results confirm that HyperSF can significantly improve the average local conductance compared with other hypergraph coarsening methods for all test cases.

\subsection{Cut preservation after hypergraph coarsening}
\begin{figure}[]
    \centering
    \includegraphics [  width=0.45\textwidth]{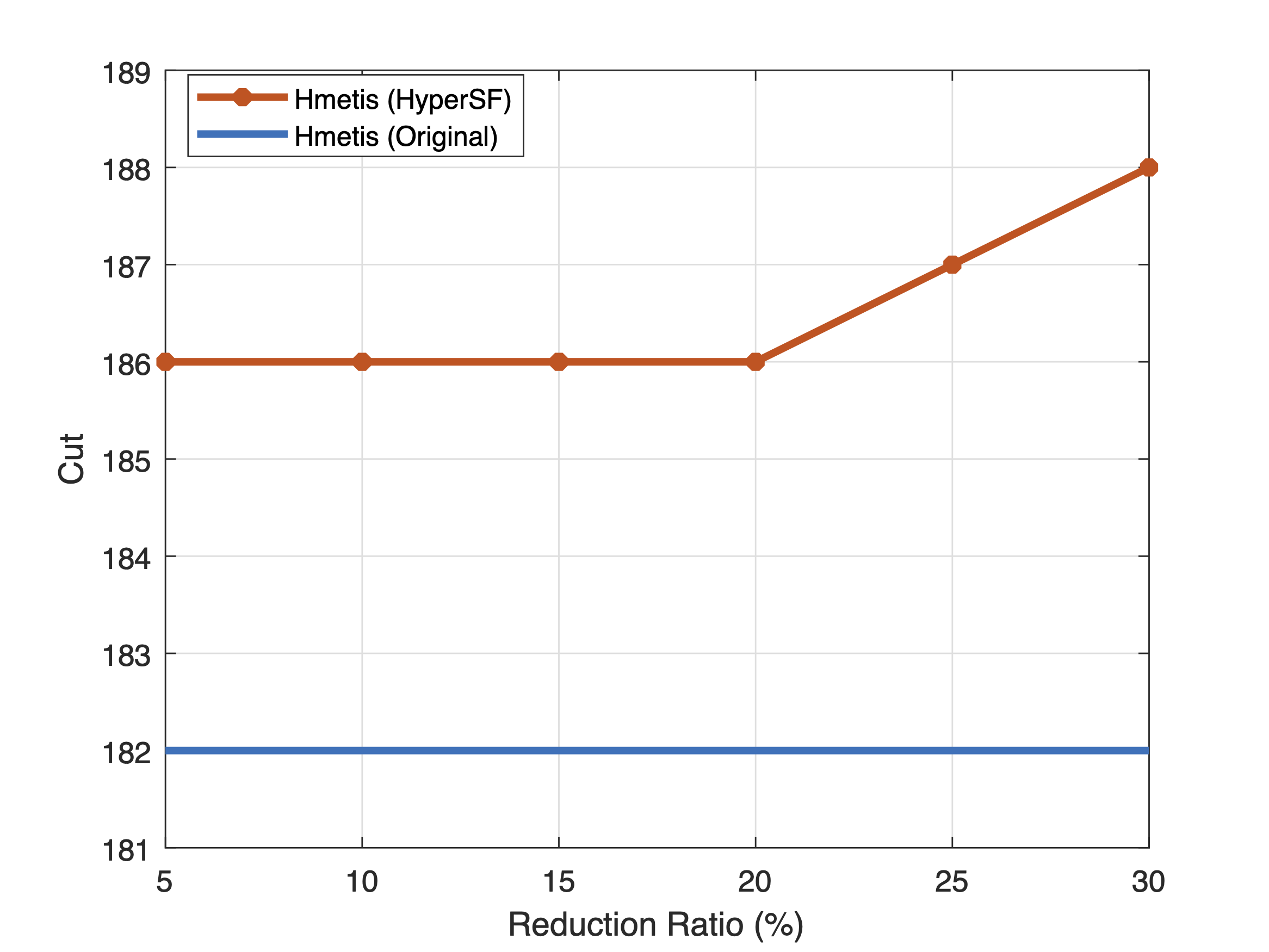}
    \caption{Cut preservation  after hypergraph coarsening}
    \label{fig:cut}
\end{figure}

To further evaluate the performance of the proposed hypergraph coarsening algorithm, we calculate the cut before and after coarsening. \textbf{Before coarsening:} Hmetis bisects the original hypergraph into two partitions and returns the number of hyperedges intersecting between two partitions; \textbf{after coarsening:} HyperSF first produces a smaller hypergraph  that is used as the input of Hmetis; then  the node partitioning results are mapped back to the original hypergraph for computing the cut. Fig. \ref{fig:cut} shows the cuts before and after applying HyperSF using various reduction ratios (RRs) for the dataset ``ibm01''. While Hmetis bisects the original hypergraph by cutting $182$ hyperedges,  only $2\%$-$3\%$ relative difference in cut is observed when   bisecting the coarsened hypergraph, which implies the coarsened hypergraphs obtained using HyperSF can very well preserve the original    cut.

\subsection{Hypergraph k-way partitioning}
The goal of this experiment is to study the performance of the proposed algorithm for k-way hypergraph partitioning tasks. We use HyperSF as a standalone hypergraph partitioner and Hmetis as the baseline to evaluate the performance and efficiency.
HyperSF aims to aggregate the spectrally-similar vertices to preserve the structural properties of the original hypergraph by ignoring the criteria heavily used in modern balanced partitioning methods. For example, Hmetis iteratively bisects the hypergraph and minimizes the hypergraph cut for achieving a  balanced partitioning result. To achieve a fair comparison of   hypergraph partitioning between HyperSF and Hmetis, we consider an imbalance parameter (\textit{UBfactor}) that is defined  to be the upper bound of the ratio between the maximum volume and the average volume. For instance, given \textit{UBfactor}$= 9$ and $Nparts = 4$ for partitioning a hypergraph with $n$ vertices, the partitioner will produce a set of partitions so that the ratio between the maximum volume and the average volume will be bounded by $1.09 \times n / 4$. We restrict the size of each partition by including \textit{UBfactor} to HyperSF, which will effectively set a limit on the size of each partition: if a vertex intends to join a partition with full capacity, it will be assigned to the nearest partition by measuring the Euclidean distance between that node and its neighborhood partition. However, adding \textit{UBfactor} will inevitably impact the solution quality due and lead to increased \textbf{HLC}. 

\begin{figure}[]
    \centering
    \includegraphics [ width=0.45\textwidth]{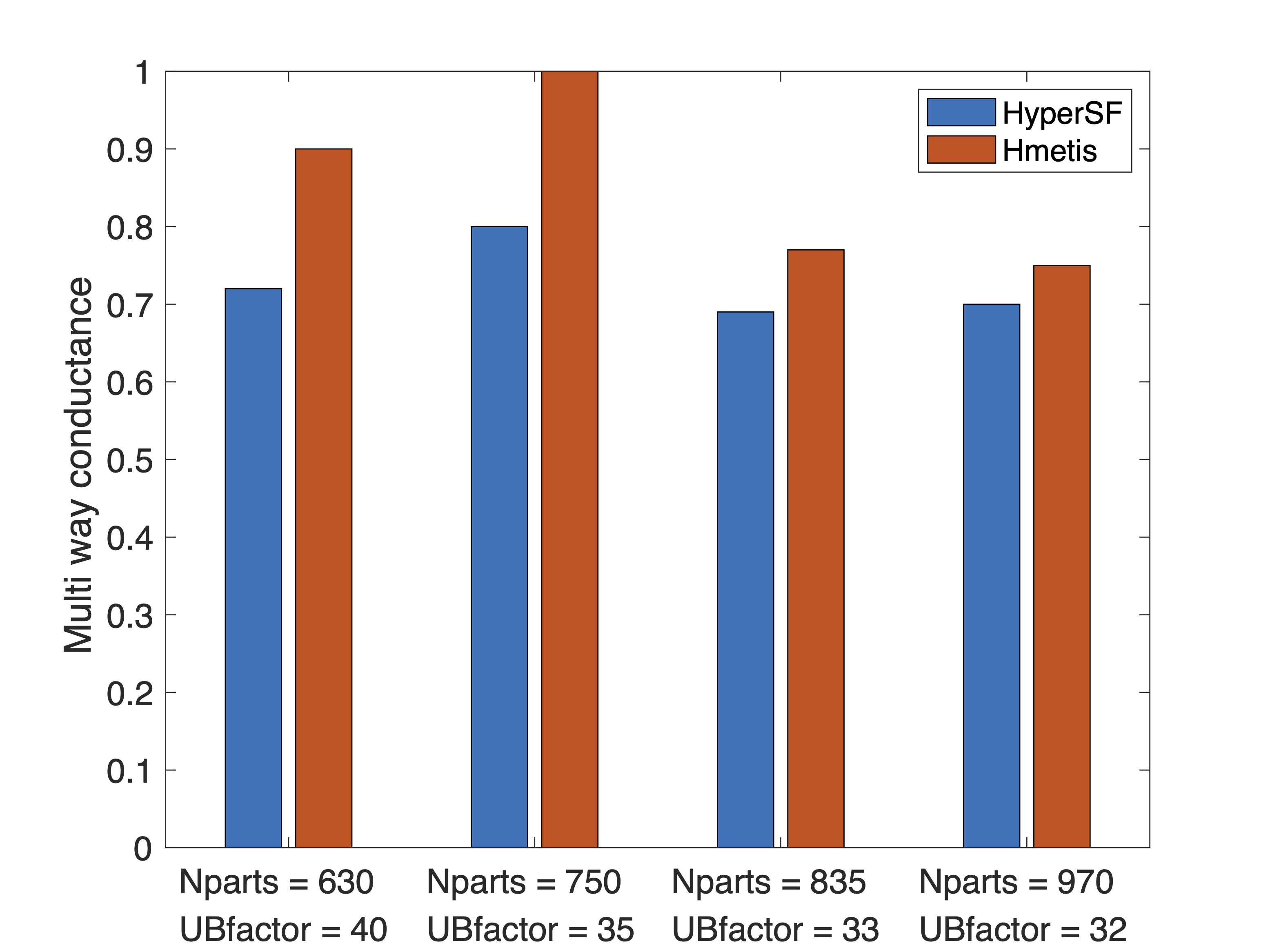}
    \caption{K-way conductance analysis for different number of partitions with respect to the imbalance parameter}
    \label{fig:multi-way conductance}
\end{figure}

Since the partitions are no longer strictly balanced, the cut objective will not be meaningful for assessing the   performance. Consequently, we use the $k$-way conductance as the metric to compare the performance of both methods. The $k$-way conductance can be computed by finding  the maximum (node cluster) conductance over all partitions \cite{lee2014multiway}. 
Fig. \ref{fig:multi-way conductance} shows the result of conducting a hypergraph $k$-way partitioning  on the ``ibm01" dataset to compute the $k$-way conductance with Hmetis and HyperSF. Both hypergraph partitioners split the vertices into $Nparts = \left[630, 750, 835, 970\right]$ through multiple experiments and imposed the same imbalance factor between the partitions. We observe a smaller $k$-way conductance when using HyperSF compared to Hmetis for various partitioning and imbalance parameter settings.   

\begin{figure}[]
    \centering
    \includegraphics [ width=0.45 \textwidth]{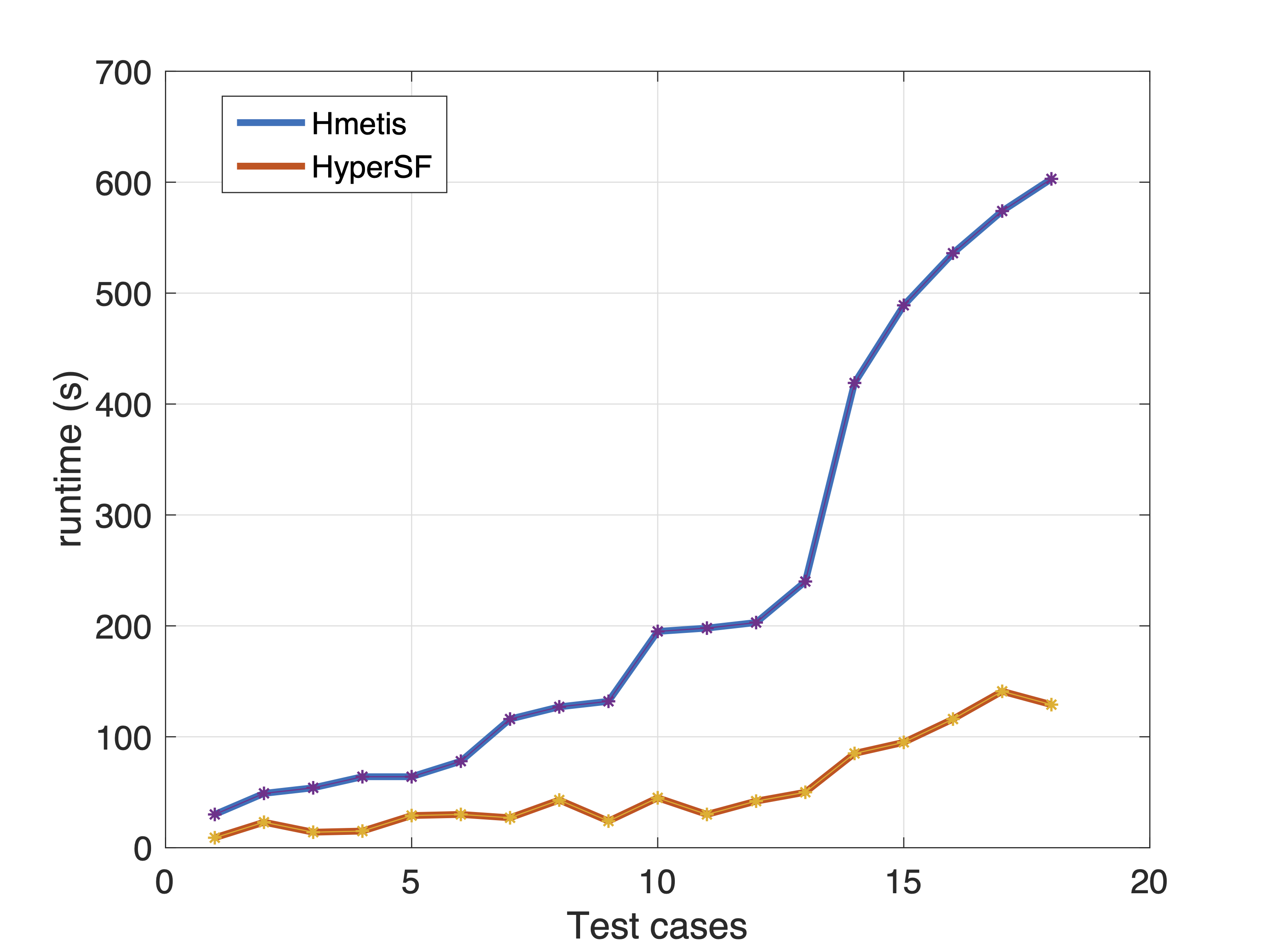}
    \caption{Runtime comparison for all test cases }
    \label{fig:run_time_curve}
\end{figure}

\begin{figure}[]
    \centering
    \includegraphics [  width=0.45\textwidth]{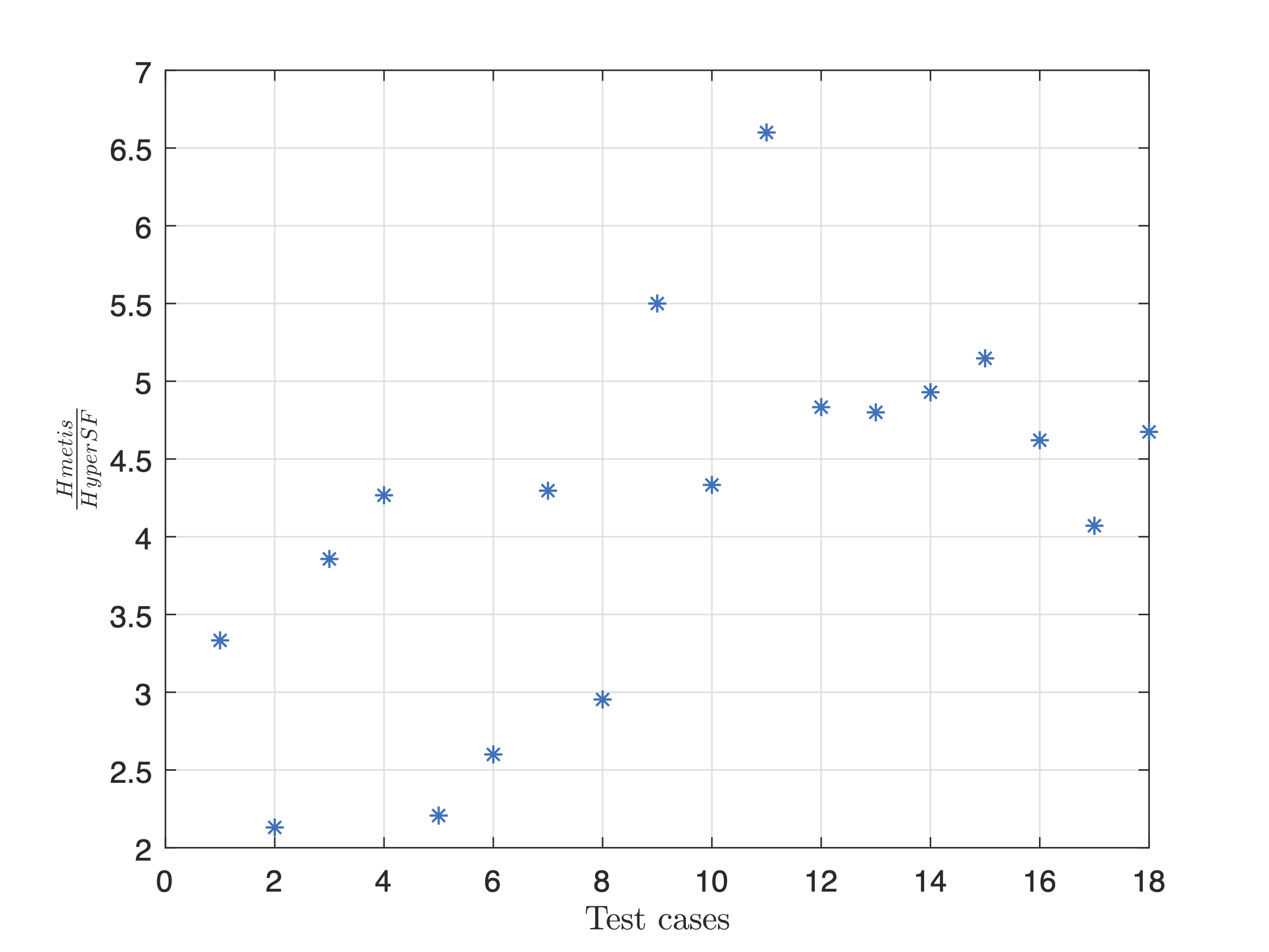}
    \caption{Runtime speedups for all test cases}
    \label{fig:run_time}
\end{figure}
\subsection{Runtime analysis}
We compared the runtime of HyperSF with Hmetis for all the test cases to evaluate the algorithm scalability. As shown in Fig. \ref{fig:run_time_curve}, the gap between the Hmetis and HyperSF keeps increasing as the hypergraph sizes increase.   Fig. \ref{fig:run_time} shows up to $6.6\times$ runtime speedup   achieved by HyperSF, which further highlights the superior runtime efficiency of HyperSF over Hmetis.

\vspace{-5pt}
\section{Conclusion} \label{sec:conclusion}
In this work, we propose  an efficient spectral hypergraph coarsening algorithm (HyperSF) for aggressively reducing the size of hypergraphs without impacting the key  spectral (structural) properties.  Our approach  leverages an initial spectral clustering procedure and a  flow-based local clustering scheme for detecting the sets of strongly-coupled hypergraph vertices.  Our  results for a variety of hypergraphs extracted from real-world VLSI design benchmarks show that the proposed HyperSF can significantly improve the solution quality and runtime efficiency   when compared with prior state-of-the-art hypergraph coarsening algorithms.  
\section*{Acknowledgment}
This work is supported in part by the National Science Foundation under Grants CCF-2041519 (CAREER), CCF-2021309 (SHF), and CCF-2011412 (SHF).
\bibliographystyle{IEEEtran}
\bibliography{Ref}  %

\end{document}